\documentclass{article}

\usepackage{arxiv}

\usepackage[utf8]{inputenc} 
\usepackage[T1]{fontenc}    
\usepackage{hyperref}       
\usepackage{url}            
\usepackage{booktabs}       
\usepackage{amsfonts}       
\usepackage{nicefrac}       
\usepackage{microtype}      
\usepackage{lipsum}
\usepackage{graphicx}
\usepackage{subcaption}
\usepackage[font=normalsize,labelfont=bf]{caption}
\usepackage{multirow}

\title{Collaborative Mapping of Archaeological Sites using multiple UAVs\thanks{This work was supported by the M N Faruqui Innovation center at the Indian Institute of Technology Kharagpur, India}}

\author{
 Manthan Patel \\
  Department of Mechanical Engineering\\
  Indian Institute of Technology Kharagpur\\
  West Bengal, India \\
  \texttt{manthan@iitkgp.ac.in} \\
   \And
 Aditya Bandopadhyay \\
  Department of Mechanical Engineering\\
  Indian Institute of Technology Kharagpur\\
  West Bengal, India \\
  \texttt{aditya@mech.iitkgp.ac.in} \\
  \And
 Aamir Ahmad \\
  Institute for Flight Mechanics and Controls\\
  The Faculty of Aerospace Engineering and Geodesy\\
  University of Stuttgart\\
  Stuttgart, Germany\\
  \texttt{aamir.ahmad@ifr.uni-stuttgart.de} \\
}

\date{}
\begin{document}
\maketitle
\begin{abstract}
UAVs have found an important application in archaeological mapping. Majority of the existing methods employ an offline method to process the data collected from an archaeological site. They are time-consuming and computationally expensive. In this paper, we present a multi-UAV approach for faster mapping of archaeological sites. Employing a team of UAVs not only reduces the mapping time by distribution of coverage area, but also improves the map accuracy by exchange of information. Through extensive experiments in a realistic simulation (AirSim), we demonstrate the advantages of using a collaborative mapping approach. We then create the first 3D map of the Sadra Fort, a 15th Century Fort located in Gujarat, India using our proposed method. Additionally, we present two novel archaeological datasets recorded in both simulation and real-world to facilitate research on collaborative archaeological mapping. For the benefit of the community, we make the AirSim simulation environment, as well as the datasets publicly available.\footnote{Project web page: http://patelmanthan.in/castle-ruins-airsim/}

\end{abstract}


\begin{figure}[h!]
     \centering
     \begin{subfigure}[b]{0.61\textwidth}
         \centering
         \includegraphics[width=\textwidth]{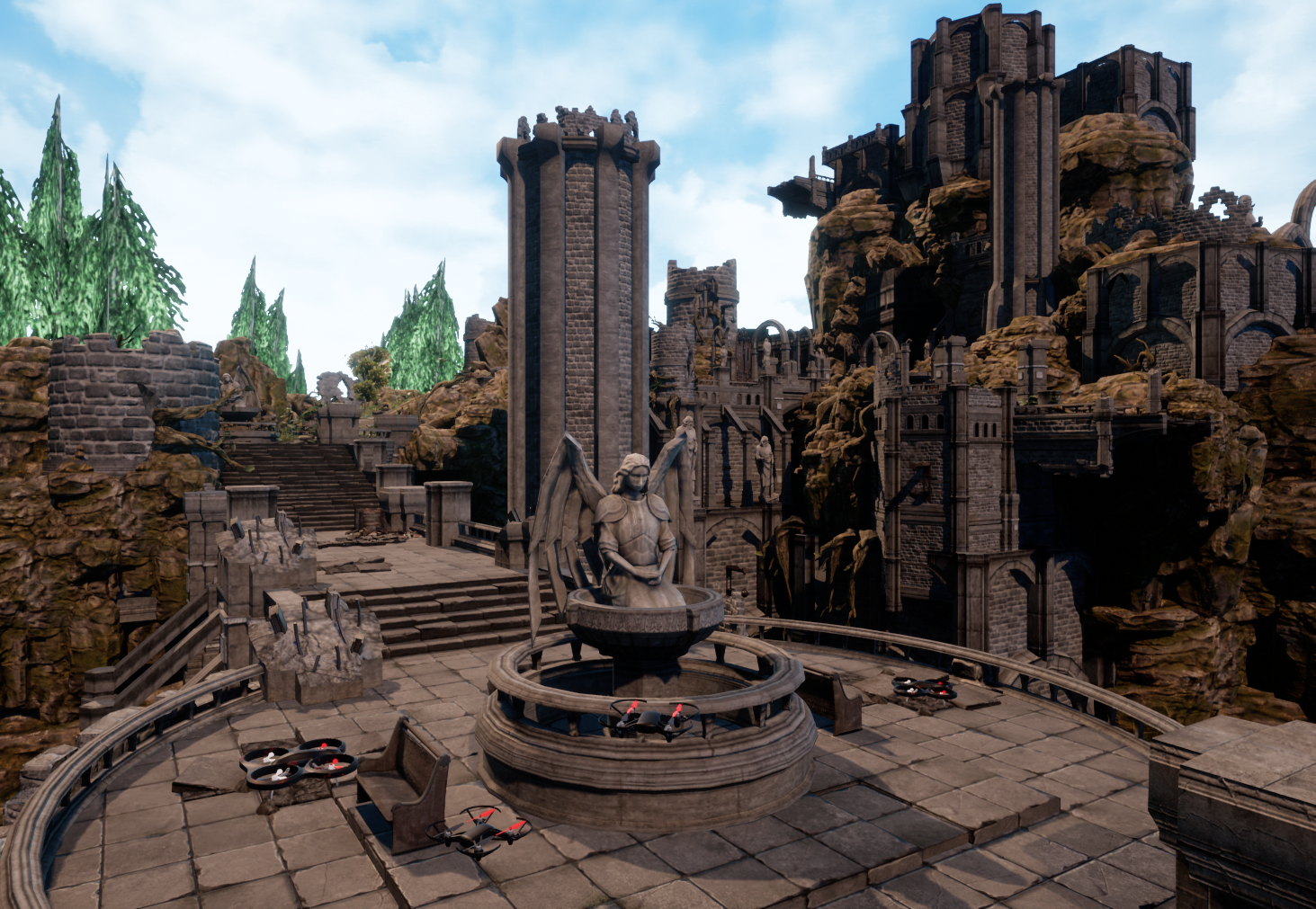}
        \caption{}
     \end{subfigure}
     \hfill
     \begin{subfigure}[b]{0.38\textwidth}
     \begin{subfigure}[b]{\textwidth}
         \centering
         \includegraphics[width=\textwidth]{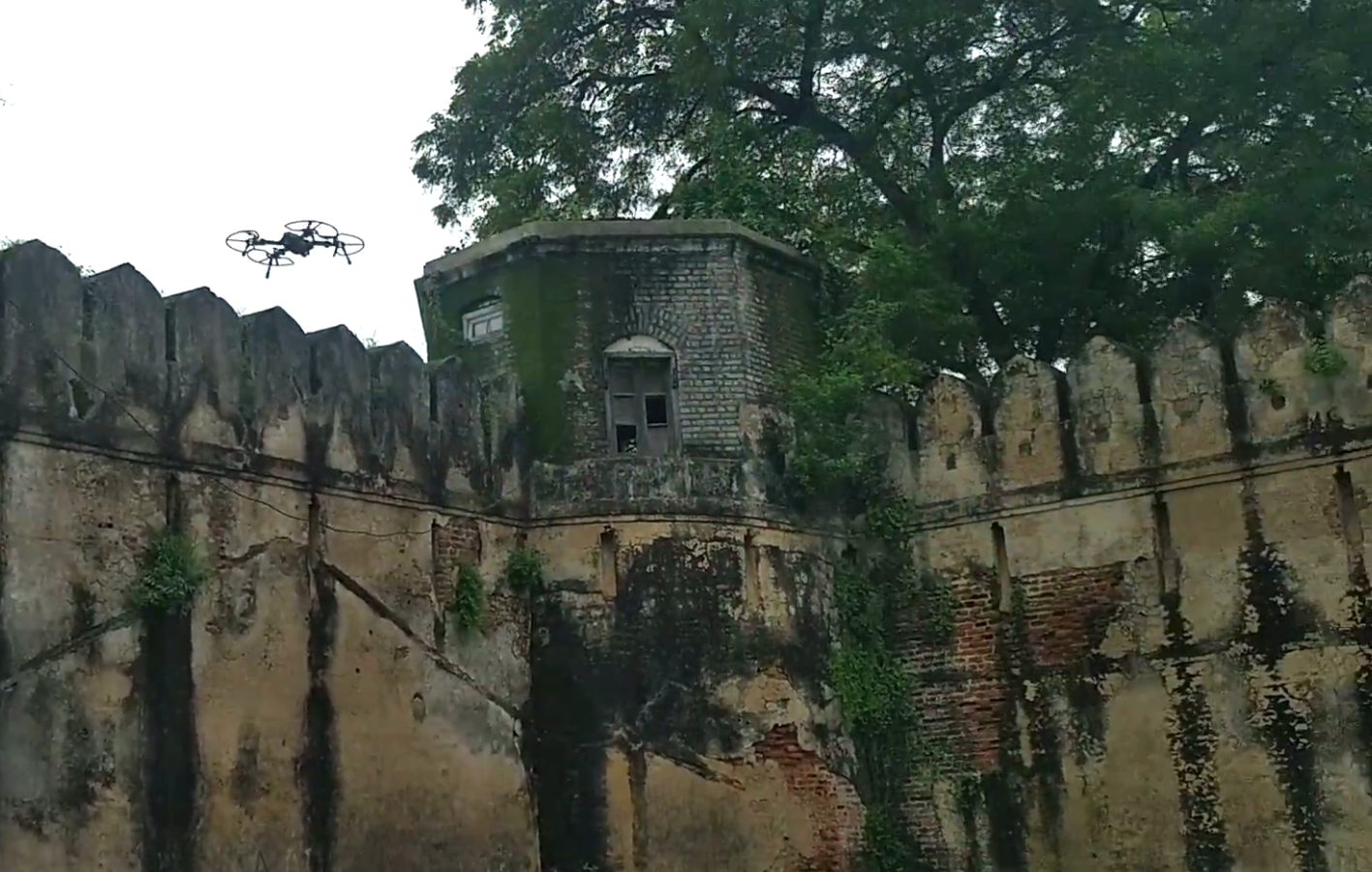}
        \end{subfigure}
         \begin{subfigure}[b]{\textwidth}
         \centering
         \includegraphics[width=0.7\textwidth]{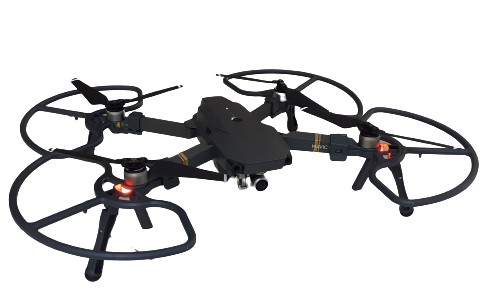}
        \caption{}
        \end{subfigure}    
     \end{subfigure}
        \caption{(a) A team of UAVs mapping the sim environment, (b) UAV in action at the Sadra Fort archaeological site}
        \label{fig:cover}
\end{figure}


\section{Introduction}

With the advent of ever increasing reliability and development of unmanned aerial vehicles (UAVs), their use for varied applications like search-and-rescue missions, reconnaissance, precision agriculture and industrial inspection etc. have been greatly increasing. One such important application is its use for archaeological reconstruction and 3D mapping of heritage monuments. Archaeologists have always been challenged by factors such as lack of accessibility and insufficient viewpoints for carrying out the study. As an example, most of the castles and forts have a significant height and it is not possible to completely map such sites using traditional hand held devices and cameras. Thus to remove these difficulties and to augment the quality of maps, attempts have been made as early as in 2009 \cite{3dvirtualreconstruction} wherein data collected by multiple sensors (lidars and cameras) mounted on a UAV were used for 3D reconstruction of complex architectures like castles. \par

UAVs can be highly stable, agile and can easily access any location. Due to these reasons, they are the preferred choice for carrying out mapping operations. However, an accurate localization is needed to ensure generation of good quality maps. Most of the UAVs today are equipped with a GPS for positioning, but due to their inaccuracies and lack of GPS signal in certain remote and subterranean locations they cannot be considered as as a reliable mode for localization. Thus, a simultaneous localization and mapping (SLAM) approach must be used to carry out the localization and mapping operation. We propose to use a visual SLAM (vSLAM) approach due to the advantages of portability, low cost and power. Light-weight cameras can be easily mounted on a UAV as opposed to the lidars which are both expensive and heavy. The ground-breaking work in ORB SLAM \cite{murORB2} has become the current gold standard for visual SLAM. But when it comes to large scale mapping, these algorithms involving a single agent often tend to drift which is difficult to correct even using loop closures and global bundle adjustment (GBA) steps. Thus it is advantageous to use a collaborative mapping strategy in this scenario to increase the map accuracy. By using multiple agents, each UAV needs to cover a small region which will lead to faster mapping of the site and less amount of drift which can be easily corrected by loop closures. The centralized collaborative monocular SLAM (CCM SLAM) approach developed in \cite{schmuck2017ccm}\cite{schmuck2017multi} is the current state of the art in collaborative visual SLAM. We build upon the aforementioned CCM SLAM approach for our work on mapping of archaeological sites.
\par

We use the Microsoft AirSim \cite{airsim2017fsr}, a high fidelity simulator based on the Unreal Engine (UE) for our experiments in simulation. We develop our own realistic archaeological environment portraying a castle ruins scene and demonstrate collaborative mapping of this archaeological site. We then perform extensive testing of the collaborative SLAM approach for up to four agents (UAVs) in such a medium to large scale environment. As an additional experiment, we carry out the collaborative sparse mapping followed by dense reconstruction of a 15th-century fort located in Gujarat, India. In summary, we make the following major contributions-
\begin{itemize}
    \itemsep-0.2em
    \item A highly realistic simulation world in UE depicting a castle ruins site that can be used for 3D archaeological mapping.
    \item The first collaborative SLAM dataset consisting of thirteen sequences, specially tuned and prepared for testing the effectiveness of collaborative SLAM algorithms. These combined sequences can be used for the full archaeological reconstruction of the castle ruins site in simulation.
    \item A rigorous evaluation of the collaborative SLAM approach using up to four agents and empirically showing its advantages over single-agent SLAM methods. 
    \item The Sadra-Fort dataset consisting of four sequences and creating the first 3D map of this real-world site.
\end{itemize}

\section{Related Work}

\subsection{Archaeological Mapping using UAVs}
The work presented in \cite{drones_archaeology} reviews the use of drones in archaeology and discusses the current state-of-the-art and future perspectives. In the earlier days, instruments such as kites, balloons, and blimps mounted with a camera were used to capture photographs of archaeological sites. With the advent of light-weight and stable UAVs, its use for archaeological mapping became popular \cite{3dvirtualreconstruction}. UAVs with a varied sensor setup can be used for collecting the data. For example, the use of lidar for archaeological mapping is discussed in \cite{lidar-uav-arch}, while the work presented in \cite{venetian-walls} used data collected by a UAV mounted with a monocular camera for carrying out the complete mapping of the Venetian walls of Bergamo. Here, the scaling of the model using terrestrial control points had to be performed due to the use of only a monocular camera. A combination of terrestrial laser scanning (TLS) and image-based techniques was used to generate a complete 3D model of a Roman Church in \cite{church-mapping}. \par

In \cite{venetian-walls}, a standard approach is developed for carrying out archaeological mapping. First, the site to be mapped is surveyed, and accordingly, a flight plan for the UAV is decided. Next, the images are captured using a UAV, and a sparse map is built using an offline structure from motion (SFM) approach. The sparse point cloud is then used to create a dense point cloud that is scaled to real dimensions according to the ground control points. As a final step in reconstruction, the dense point cloud is converted to a mesh followed by texture projection on the mesh. The 3DFlow Zephyr software is used to carry out the above steps for model creation. 

\cite{opensource-recon-pipelines} reviews the open-source image-based 3D reconstruction pipelines, which can be used for archaeological mapping. A comparative study between three of the commonly available open-source pipelines, namely OpenMVG+OpenMVS \cite{moulon2016openmvg}\cite{openmvs2020}, COLMAP \cite{schoenberger2016mvs} and AliceVision\cite{Alice1-Moulon2012} is performed. Each of these approaches solves an SFM problem to generate a sparse map, which is then densified and converted into a mesh using multi-view stereo (MVS) approaches. A critical evaluation of the three approaches is carried out in diverse scenarios, and a comparable performance is observed, with the COLMAP and OpenMVS slightly outperforming the AliceVision framework.

\subsection{Collaborative SLAM}

A collaborative monocular SLAM system with a focus on dynamic environments is presented in coSLAM \cite{coslam}. In their approach, image streams from multiple monocular cameras are taken as input, and the cameras are grouped according to the overlap in their views. Inter-camera pose estimation and inter-camera mapping are introduced to deal with dynamic objects in the localization and mapping process. But, since the cameras need to be synchronized in coSLAM, it is not a convenient method for mobile robots, wherein the communication between robots may contain high latency and may not be reliable.  \par

With an objective to allow multiple clients to share and merge maps, MOAR SLAM was developed in \cite{MoarSLAM}. Here, each agent runs a full onboard SLAM, including the computationally heavy tasks of place recognition and global optimizations. The server is used for storing and sharing the client's maps. Though the server performs the map matching, it does not perform the global optimization, and thus the agents require a high computation power to carry out these expensive tasks. In \cite{back-coslam}, a framework for real-time multi-robot collaborative SLAM is presented. The framework combines local pose graphs obtained from its multiple robot-agents into a global pose graph, which it then feeds back to the robots to increase their mapping and localization effectiveness. But, this framework requires the pose graph along with the associated image for each Key Frame and its absolute scale as an input. Since most of the Monocular graph-based SLAM methods like ORB SLAM \cite{murTRO2015} and PTAM \cite{PTAM} do not provide an absolute scale, it is not possible to use this framework \cite{back-coslam} with these approaches. \par

The CCM SLAM approach is developed in \cite{schmuck2017multi} and \cite{schmuck2017ccm} which is the current state-of-the-art in collaborative monocular SLAM. Here, each of the agents perform the tasks of tracking and local mapping onboard, while the computationally expensive tasks of place recognition, loop closure detection, map fusion, and global optimization are performed by the server.

\section{Methodology}
The existing methods for archaeological reconstruction employ an offline SFM method to construct a map which is both time consuming and computationally expensive. We thus propose a two-step method for archaeological reconstruction which is must faster and involves the following steps.
\begin{itemize}
    \item Step 1: A collaborative SLAM approach (Section \ref{sec: CCM SLAM architecture} and \ref{sec:results}) for developing a sparse map using UAV agents. This step can be performed online and is much faster than offline SFM approaches.
    \item Step 2: An offline Multi-view Stereo (MVS) approach for dense reconstruction using the sparse map developed in step 1. This step is discussed in Section \ref{sec:expt5}.
\end{itemize}
 
\subsection{CCM-SLAM Architecture}
\label{sec: CCM SLAM architecture}
\begin{figure}[!ht]
    \centering
    \includegraphics[width=1\textwidth]{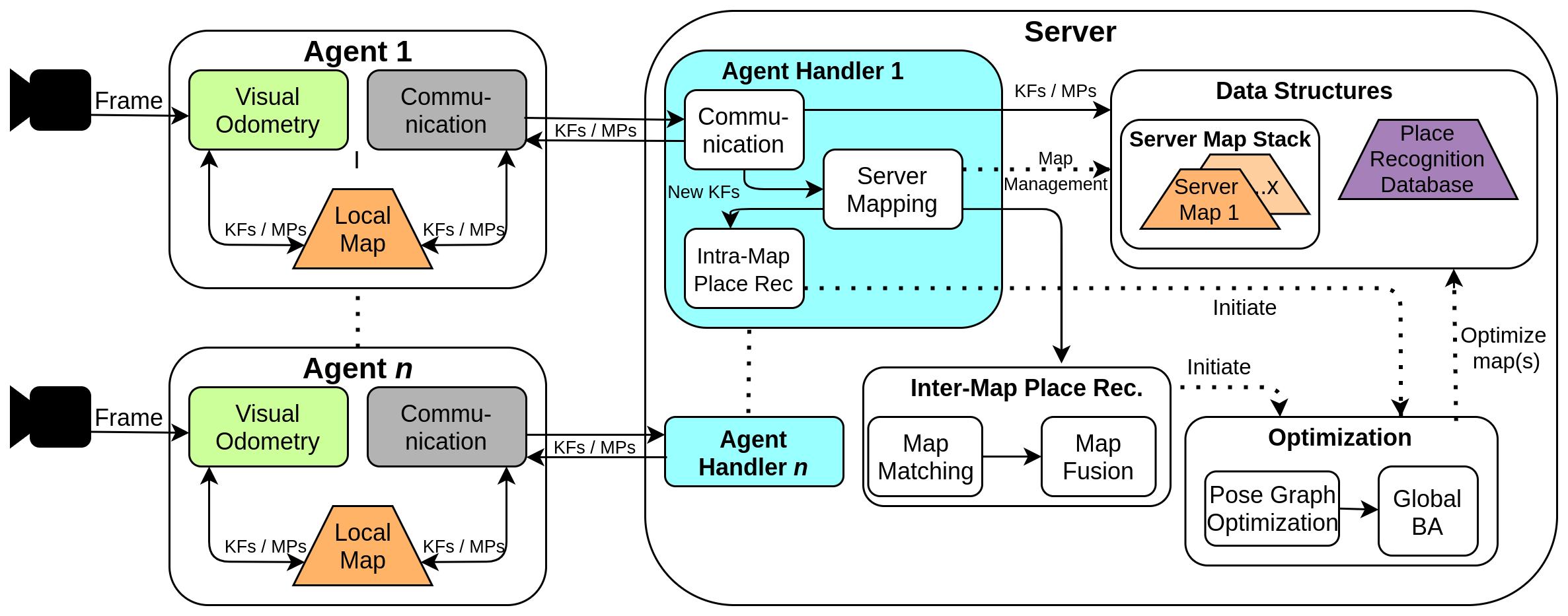}
    \caption{The overview of the CCM-SLAM architecture}
    \label{fig: CCM SLAM architecture}    
\end{figure}

 We implement and build upon the CCM SLAM \cite{schmuck2017ccm}. We use ROS \cite{ROS} as our base framework. CCM SLAM is designed for multiple UAV's each having a monocular camera and small computation capabilities. In addition to the agents, there is a centralized ground station (server) that has a relatively higher computation power and handles all the expensive tasks for each of the agents. There is a bi-directional information exchange between the server and the agents. The framework has been designed in such a way that the individual agents always preserve their autonomy by running the critical tasks for navigation on-board while outsourcing the computationally expensive tasks such as place recognition and Global Bundle Adjustments (GBA) to the server. Each of the agents are initialized separately and function independently until a map fusion operation is carried out (which occurs when a significant overlap is found between maps of different agents). A robust communication strategy has been developed, which can handle network delays, message loss, and limited bandwidth, which frequently occur in a real-world setup. The overview of the system architecture can be seen in Figure \ref{fig: CCM SLAM architecture}. The server maintains an agent handler for each of the agents and performs the computationally expensive tasks as described above. 

\subsection{The Agents}
We use a Key Frame (KF) based visual odometry (VO) developed in \cite{murTRO2015} for each of the n agents. Each agent has three processes - tracking, mapping, and communication, which run in parallel threads. The tracking thread estimates the frame-to-frame movement of the camera and optimizes the pose using motion-only BA. In addition to this, it also decides whether the incoming frame is eligible for being a KF. The mapping thread processes new KFs and performs a local BA to optimize the local map surrounding the KF. The local map L is a SLAM graph where the KFs and Map points (MPs) are connected by edges. Two KFs are connected by an edge if they share a minimum number of common visible MPs. The weight corresponding to this edge depends on the number of shared observations between them. This local map L is restricted to N KFs to limit the onboard computation load. When the number of KFs in map L exceeds N, a trimming operation is performed to remove the older KFs and keeping the most recent N KFs. The communication module on the agent's side is responsible for sending the added and changed KFs and MPs to the server. The server returns the KFs and their observed MPs nearest to (having maximum covisibilty) the previous KF of the agent. Having this additional information that may have been acquired by some other agent while visiting the same place augments the local map, which helps in better pose estimation of the agent. \par

\subsection{The Server}
On the server-side, there is a separate agent handler for each of the agents, and a server map is maintained for each of the agent. The agent handler keeps a Sim(3)-transformation that transforms the data from the local map of the agent to the server map. This is especially important when there is a map fusion of agents having different reference frames. When a map fusion occurs, the server maps of the respective agents are replaced by a third map consisting of information from both the agent's server maps, followed by a global optimization step. For example, if a map fusion of agents 0 and 1 occur, a third new server map is created which contains the fused map of agents 0 and 1, after which their individual server maps get deleted. Any new information from either of the agents is now added to this newly formed server map. Thus if there are four agents, at any time, the number of maps in the server range from one to four (a single map will exist if all four maps have been fused). \par
The KF database used for place recognition is built incrementally using all acquired KFs from each agent. A hierarchical bag of words approach \cite{DBoW2} is used for fast and efficient place recognition. Two types of place recognition queries - intra-map place recognition and map matching  (inter-map) are performed. These help in adding new constraints to the pose-graph when revisiting a place. Successful loop detection is followed by a loop closure, essential graph optimization \cite{murTRO2015}, and a GBA step. The GBA optimizes the server map by minimizing the reprojection error for all the KFs and MPs, to improve the accuracy of the map and reduce the drift in scale which occurs for larger trajectories. The Levenberg-Marquardt algorithm of g2o \cite{g2o} is used for GBA. 

\section{Environment Setup and Dataset}

We perform experiments in simulation as well as on a real-world archaeological site. We use the Microsoft AirSim \cite{airsim2017fsr} for our simulation work due to its highly realistic graphics, easy integrability, and support for multiple UAVs and sensor setups. Despite providing a variety of environments like that of a city, neighborhood, and forest, at present, no environment depicting an archaeological site is available for AirSim, hence we develop our own environment, the Castle Ruins (CR) AirSim, which has been discussed in Section \ref{sec:CR_airsim}. We create our own monocular camera dataset, the CR-Monocular dataset (Section \ref{sec:CR-mono-dataset}), which is specially recorded and tuned to be used for evaluating collaborative SLAM approaches. The UAV setup for the experiments performed at the real-world archaeological site, Sadra Fort (SF) along with the its dataset preparation is presented in Section \ref{sec:SF-setup-dataset}. For the benefit of the community, we make the CR-AirSim environment, along with the CR and SF dataset, publicly available.  

\subsection{CR AirSim Environment}
\label{sec:CR_airsim}

AirSim is an open-source, cross-platform simulator for drones and ground vehicles, built on Unreal Engine 4 (UE4). UE4 is best known for its photo-realistic visuals, and thus, it can emulate highly realistic scenes. AirSim comes with a default quadrotor UAV model of the AR-Drone. It is modeled using a physics engine for linear and angular drag, while the environment model is made up of gravity, magnetic field, air pressure, and density models. Moreover, the UAV model also consists of a sensor model for the barometer, magnetometer, gyroscope, accelerometer, and GPS. Different camera models such as the RGB, RGB-D, stereo, and IR-camera are available for use, and a UAV can be mounted with a maximum of five cameras. The models are created to facilitate a real-time flight. Apart from the default AR-Drone, we develop a new model for the DJI Tello quadrotor. Due to its smaller size (l*b*h:98mm*92.5mm*41mm), it can easily navigate in constrained indoor environments. We first create a mesh for the Tello in Solidworks which is then imported to the UE4 editor. While the basic quadrotor physics model remains the same, we tweak the dimensions, thrust and weight parameters to develop the Tello model. \par

We develop the CR AirSim environment, which emulates a highly realistic archaeological site depicting a ruins scene. Spread across an area of 300x300 meters, the environment consists of diverse features and places ranging from sparse features to feature-rich locations. We make use of the freely available content resources of the game Infinity Blade for developing the environment. Due to its setting in a medieval-age scenario, plenty of resources such as castles, statues, and other ruins architecture are available for use. We assemble these resources to develop a castle-ruins scene and make it compatible with AirSim. We add the feature for dynamic lighting and weather conditions that can be set according to the requirements and changes with the day's simulation time. Multiple UAVs can be deployed in this environment. Various sensors and cameras can be attached to these UAVs, as discussed above. To our knowledge, no such open-source AirSim environment of an archaeological site exists, which can be extensively used for SFM, SLAM, and even exploration-mapping.


\subsection{CR Monocular cam dataset}
\label{sec:CR-mono-dataset}

We record a monocular camera dataset in the above described CR AirSim environment, which we use for our testing and evaluation. To our knowledge, no such dataset exists, which has been recorded for performing evaluation of collaborative SLAM approaches. The dataset consists of sequences that depict diverse scenarios. For example, some of the sequences have a lot of visibility in common, which can be used to evaluate the amount of improvement in trajectory prediction with an increase in covisibilty among different agents. On the other hand, some of the sequences contain only a small number of common places, which can be used to test the robustness of the place recognition module. Moreover, the dataset covers an entire archaeological site that can be used to demonstrate the mapping and reconstruction of such a site. The dataset comprises of 13 different sequences, each recorded at 640x480p (monochrome) resolution at 20 Hz using a front-facing UAV mounted camera. The camera's intrinsic and extrinsic parameters, along with the ground truth (AirSim provides the ground truth for each UAV), have been provided for each of the trajectories. We provide the camera calibration YAML file and the trajectory ground truth in the format of the Odometry ROS message type bag file as well as in the CSV format. \par
\begin{figure}
\setlength{\belowcaptionskip}{-5pt}
     \centering
     \begin{subfigure}[b]{0.505\linewidth}
         \centering
         \includegraphics[width=\textwidth]{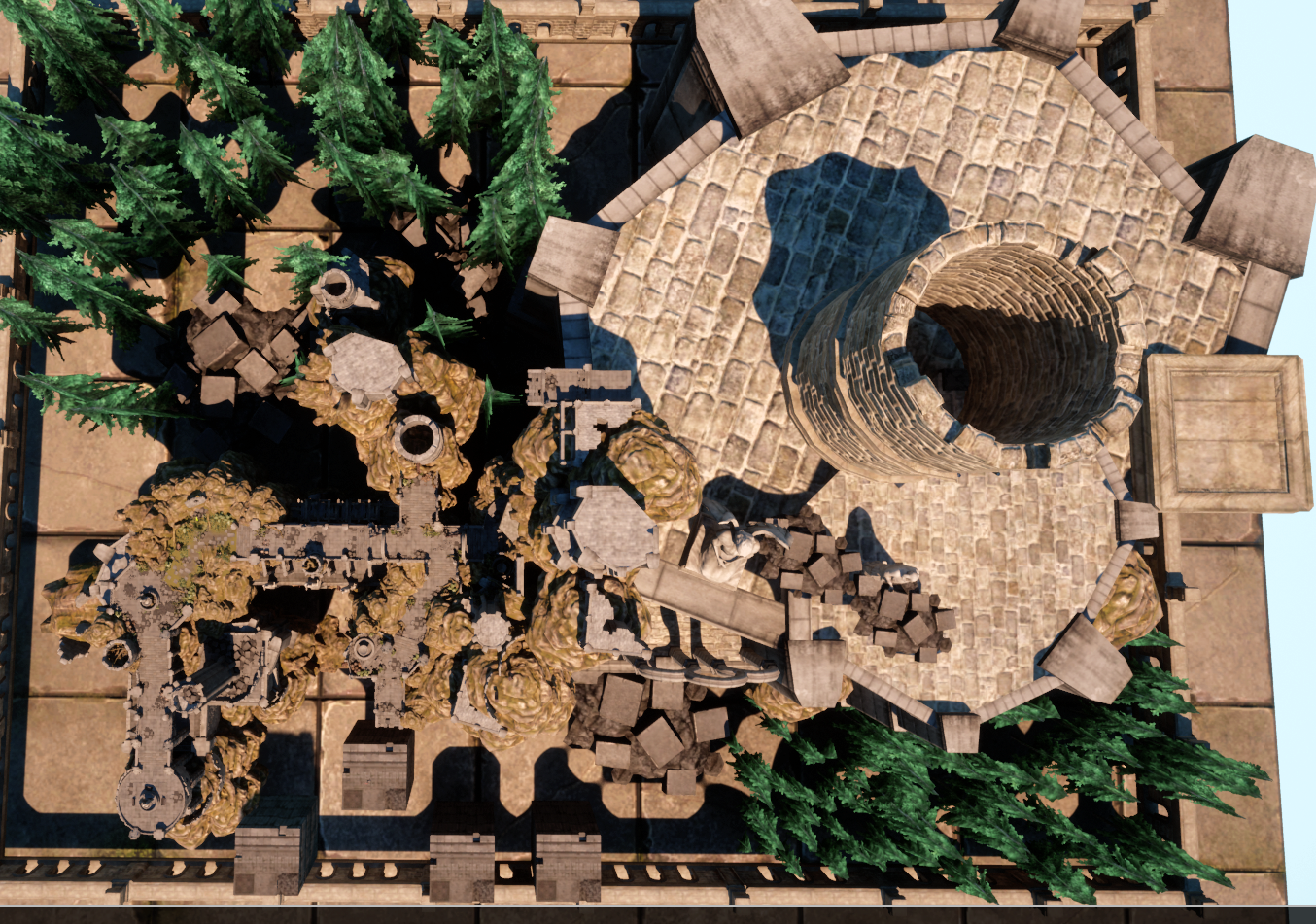}
         \caption{}
         \label{fig:CR_top}
     \end{subfigure}
     \hfill
     \begin{subfigure}[b]{0.48\linewidth}
         \centering
         \includegraphics[width=\textwidth]{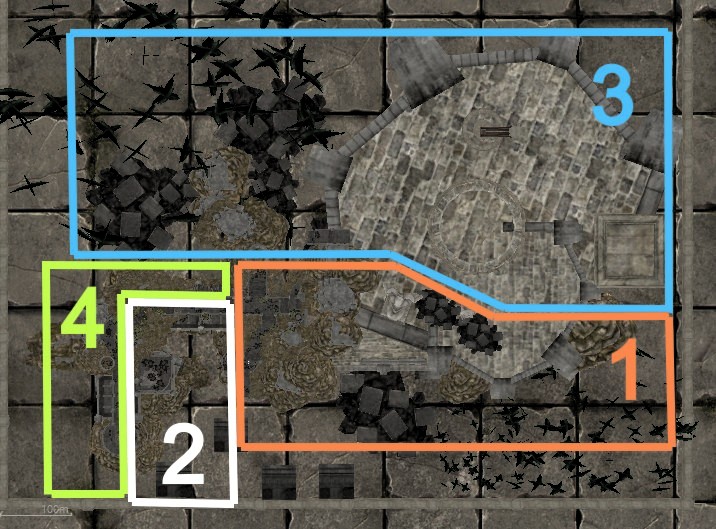}
         \caption{}
         \label{fig:CR_zones}
     \end{subfigure}
        \caption{(a) shows the top view of the CR environment and  (b) shows the division of the CR environment into 4 zones for preperation of the dataset}
        \label{fig:CR-zones}
\end{figure}

The thirteen sequences have been divided into five categories, as summarized in Table \ref{tab:dataset}. The CR-S series consists of three trajectories recorded at lower velocities and have shorter trajectory lengths. These velocities are particularly comparable to the data recorded in the EuRoC dataset \cite{EuRoC}. The sequences from the remaining four categories can be used for mapping the archaeological environment. These have been recorded at medium-high speed velocities with an average velocity ranging from 2 m/s up to 4.5 m/s, while the peak velocity going up to 10.5 m/s in some of the sequences. They also have a higher trajectory length ranging from 250 m to 1050 m, which is quite high when compared to the EuRoC dataset, which consists of sequences with less than 100 m trajectory length recorded in a small indoor environment. The CR environment has been divided into four different zones, as shown in Figure \ref{fig:CR-zones}. The sequence category number corresponds to the zone in which it has been recorded. For example, the name CR-1b implies that it is the second sequence from Zone 1. The trajectories from the same category will have a significant amount of overlap in terms of observed features and places. One trajectory from each of the categories can be used to create a full map of the CR environment. There will be a slight amount of overlap between different category trajectories at multiple places, which can be used for map fusion and inter-map loop closures. Each of the sequences have at least one loop closure except for the CR-4a sequence. This has been deliberately done to observe how the inter-map loop closure can be used to significantly reduce the error in trajectories devoid of independent loop closure.\par



We also provide a bag of words vocabulary \cite{DBoW2} that can be used for place recognition in various environments. We create this visual vocabulary in an offline step using 5K images from the city center and New College dataset \cite{CumminsNewmanIJRR08} and other 3K images taken from the AirSim CR environment and five other real world archaeological sites. We build a vocabulary of 6 levels and 10 clusters per level, getting around one million words. Such a large vocabulary is found to be efficient for place recognition in large image databases, as suggested in \cite{DBoW2}. 

\subsection{Real-world setup and the Sadra Fort (SF) dataset}
\label{sec:SF-setup-dataset}

We choose the SF archaeological site located in Gujarat, India, for our real-world experiments. Located on the banks of the river Sabarmati, the fort was built in the 15th Century and is spread across an area of more than 10000 sq. meters. The main site consists of a small palace which is covered with walls on three sides. The large area and ease of dividing the area into zones make it a perfect site for testing our collaborative SLAM approach. \par

We use a DJI Mavic Pro for preparing the SF monocular camera dataset. The Mavic Pro is a quadrotor with a 3-axis gimbal stabilization for the monocular camera. The camera can record at a resolution of up to 4096×2160 pixels at 24 frames per second (fps). Since it is time-consuming to process such large images and a good performance is obtained even at a lower resolution, we record the dataset at 1280x720 pixels at 30 fps. The SF dataset consists of four sequences, each covering a different part of the fort. While the SF-2, SF-3, and SF-4 focus on mapping the three walls, the SF-1 is recorded for mapping the palace. The characteristics of the dataset are shown in Table \ref{tab:SF-dataset}. The length of the trajectories lie between 100 to 300 meters, and the data is recorded at a relatively lower average velocity (0.5-1 m/s) to obtain stable image sequences. \par

Since the ground-truth trajectory is not available in this case, we need to rely on the GPS for an approximate estimate. The DJI Mavic Pro GPS has a vertical accuracy of $\pm$ 0.5 m and a horizontal accuracy of $\pm$ 1.5 m. Using the GPS and the roll, pitch, and yaw data from the flight record, we provide an approximate trajectory ground truth in a world coordinate frame of reference whose origin is fixed at the same latitude, longitude, and altitude for all the sequences. We provide the sequences in a ROS bag file containing the video sequence at 30fps and the approximate ground-truth pose at synchronized 10fps (lower due to the limitations of the DJI GPS update rate). In addition to this, we provide the camera calibration matrix and the distortion matrix, which is obtained using the standard checkerboard calibration technique \cite{camera-calib}. 


\begin{table}[!h]
    \begin{minipage}{.5\linewidth}
      \centering
       \begin{tabular}{ccccc}
\hline
\begin{tabular}[c]{@{}c@{}}Seq. \\ Name\end{tabular} &
  \begin{tabular}[c]{@{}c@{}}Dura-\\ tion\end{tabular} &
  \begin{tabular}[c]{@{}c@{}}Traj. \\ length\end{tabular} &
  \begin{tabular}[c]{@{}c@{}}Avg. \\ velocity\end{tabular} &
  \begin{tabular}[c]{@{}c@{}}Max\\ velocity\end{tabular} \\
      & (s) & (m)  & (m/s) & (m/s) \\ \hline
CR-S-a & 141 & 66   & 0.46  & 1.23  \\
CR-S-b & 167 & 51   & 0.3   & 1.05  \\
CR-S-c & 109 & 52   & 0.48  & 1.07  \\
CR-1a  & 135 & 510  & 3.75  & 10.09 \\
CR-1b  & 188 & 704  & 3.71  & 10.38 \\
CR-1c  & 206 & 806  & 3.91  & 10.26 \\
CR-2a  & 97  & 254  & 2.62  & 6.49  \\
CR-2b  & 137 & 394  & 2.84  & 7.23  \\
CR-2c  & 188 & 430  & 2.28  & 5.74  \\
CR-3a  & 201 & 864  & 4.28  & 10.15 \\
CR-3b  & 287 & 855  & 2.89  & 7.56  \\
CR-3c  & 291 & 1030 & 3.52  & 10.13 \\
CR-4a  & 180 & 310  & 1.72  & 5.17  \\ \hline
\end{tabular}
\vspace{2mm}
    \caption{CR Monocular dataset characteristics}
    \label{tab:dataset}
    \end{minipage}%
    \begin{minipage}{.5\linewidth}
      \centering
      \begin{tabular}{ccccc}
\hline
\begin{tabular}[c]{@{}c@{}}Seq. \\ Name\end{tabular} &
  \begin{tabular}[c]{@{}c@{}}Dura-\\ tion\end{tabular} &
  \begin{tabular}[c]{@{}c@{}}Traj. \\ length\end{tabular} &
  \begin{tabular}[c]{@{}c@{}}Avg. \\ velocity\end{tabular} &
  \begin{tabular}[c]{@{}c@{}}Max \\ velocity\end{tabular} \\
     & (s) & (m) & (m/s) & (m/s) \\ \hline
SF-1 & 370 & 198 & 0.535 & 1.24  \\
SF-2 & 326 & 223 & 0.684 & 2.1   \\
SF-3 & 399 & 296 & 0.742 & 2.83  \\
SF-4 & 200 & 117 & 0.585 & 2.42  \\ \hline
\end{tabular}
\vspace{2mm}
\caption{SF Monocular dataset characteristics}
\label{tab:SF-dataset}
    \end{minipage}

\end{table}

\section{Experiments and Evaluation}
\label{sec:results}

We perform three different experiments (Sections \ref{sec:expt1}, \ref{sec:expt2}, \ref{sec:expt3}) for the evaluation of the collaborative SLAM algorithm and two experiments to demonstrate the mapping of archaeological sites (Sections \ref{sec:expt3}, \ref{sec:expt4}). We construct a dense model of the Sadra Fort in Section \ref{sec:expt5}. The first three experiments have been performed using the CR-monocular camera dataset, while the final two experiments are performed using the SF-monocular camera dataset. We use a Core i7-6700HQ @ 2.60GHz x 4 and 16 GB RAM as the server for all the experiments while using lower computation for the agents. We perform the experiments in an offline mode by playing back the dataset sequences parallely in the ROS environment. We use the Absolute Trajectory Error (ATE) to evaluate the performance. For all evaluation, the KF poses obtained after loop closures and GBA in the server are utilized. The toolbox developed in \cite{Zhang18iros} is used to calculate the ATE. Since we are using a monocular system, it does not recover scale, and hence a Sim-3 transformation has to be applied between the ground-truth trajectory and the estimated trajectory for alignment. We make use of all the poses for aligning the trajectory for the best fit. Each of the experiments is performed three times and the mean ATE (RMSE) has been recorded.

\subsection{Single-agent evaluation on CR dataset}
\label{sec:expt1}

We evaluate the performance of a single UAV agent on each of the thirteen sequences of the CR-Monocular cam dataset  (Section \ref{sec:CR-mono-dataset}) using the developed collaborative SLAM approach. The single-agent method is essentially equivalent to the ORB-SLAM approach. We tune the parameters and keep them the same for all the experiments performed in simulation. We extract 1000 ORB features in each image and keep 8 levels in the scale pyramid with a scale factor of 1.2, as suggested in \cite{murTRO2015}. We keep the local map size N equal to 50. The results have been summarised in Table \ref{tab:expt1}. The CR-S series, which has been recorded at a lower velocity and having a shorter path length, has the least ATE out of all the sequences. This clearly shows that the ATE is bound to increase with the increase in the velocity. Moreover, the chances of the trajectory to drift in scale with time also increases with the increase in trajectory length, and hence, loop closures are essential to correct this drift. In the CR-S series, the CR-S-b is the only trajectory having a loop closure, and thus it is having a lesser ATE when compared to the other two sequences.\par
Since all the other sequences except the CR-S have been recorded at a higher velocity and path length, they have a significant amount of ATE. As expected, the number of KFs increase with the increase in the length of the trajectory. We observe a surprising result for the CR-1c sequence, which has an ATE quite less than the CR-1a and CR-1b despite having a considerably larger trajectory length. This can be justified due to the occurrence of loop closures at multiple places, unlike the other trajectories, which only have a single loop closure. The CR-3c sequence has a trajectory length of over 1000 m and thus a high ATE of 3.49 m. The high ATE in the sequence CR-4a can be explained by the non-existence of any loop closure. We also calculate the RMSE(\%) (the ATE per 100 m of path length), which is useful for comparing results relative to the trajectory length.

\begin{table}[!h]
  \centering
 \addtolength{\tabcolsep}{3pt}  
\begin{tabular}{cccccc}
\hline
Seq. Name & Kfs & RMSE   & RMSE   & Seq. time & Traj. Length \\
          &     & (m)    & (\%)   & (s)       & (m)          \\ \hline
CR-S-a    & 72  & 0.1420 & -      & 140       & 61           \\
CR-S-b    & 116 & 0.0755 & -      & 166       & 50           \\
CR-S-c    & 121 & 0.1734 & -      & 109       & 55           \\
CR-1a     & 213 & 0.9444 & 0.2951 & 90        & 320          \\
CR-1b     & 313 & 0.9861 & 0.2191 & 141       & 450          \\
CR-1c     & 415 & 0.7588 & 0.1032 & 185       & 735          \\
CR-2a     & 161 & 0.4342 & 0.2171 & 70        & 200          \\
CR-2b     & 222 & 1.1680 & 0.3893 & 98        & 300          \\
CR-2c     & 389 & 1.3885 & 0.3471 & 178       & 400          \\
CR-3a     & 392 & 2.4179 & 0.2811 & 200       & 860          \\
CR-3b     & 528 & 2.3335 & 0.2745 & 287       & 850          \\
CR-3c     & 568 & 3.4932 & 0.3493 & 280       & 1030         \\
CR-4a     & 382 & 2.1010 & 0.6770 & 179       & 310          \\ \hline
\end{tabular}
\addtolength{\tabcolsep}{-3pt}  
\vspace{2mm}
\caption{Evaluation of CR Monocular cam dataset}
  \label{tab:expt1}
\end{table}

\subsection{Collaborative SLAM with significant trajectory overlap}
\label{sec:expt2}

\begin{figure}
    \centering
    \includegraphics[width=1\textwidth]{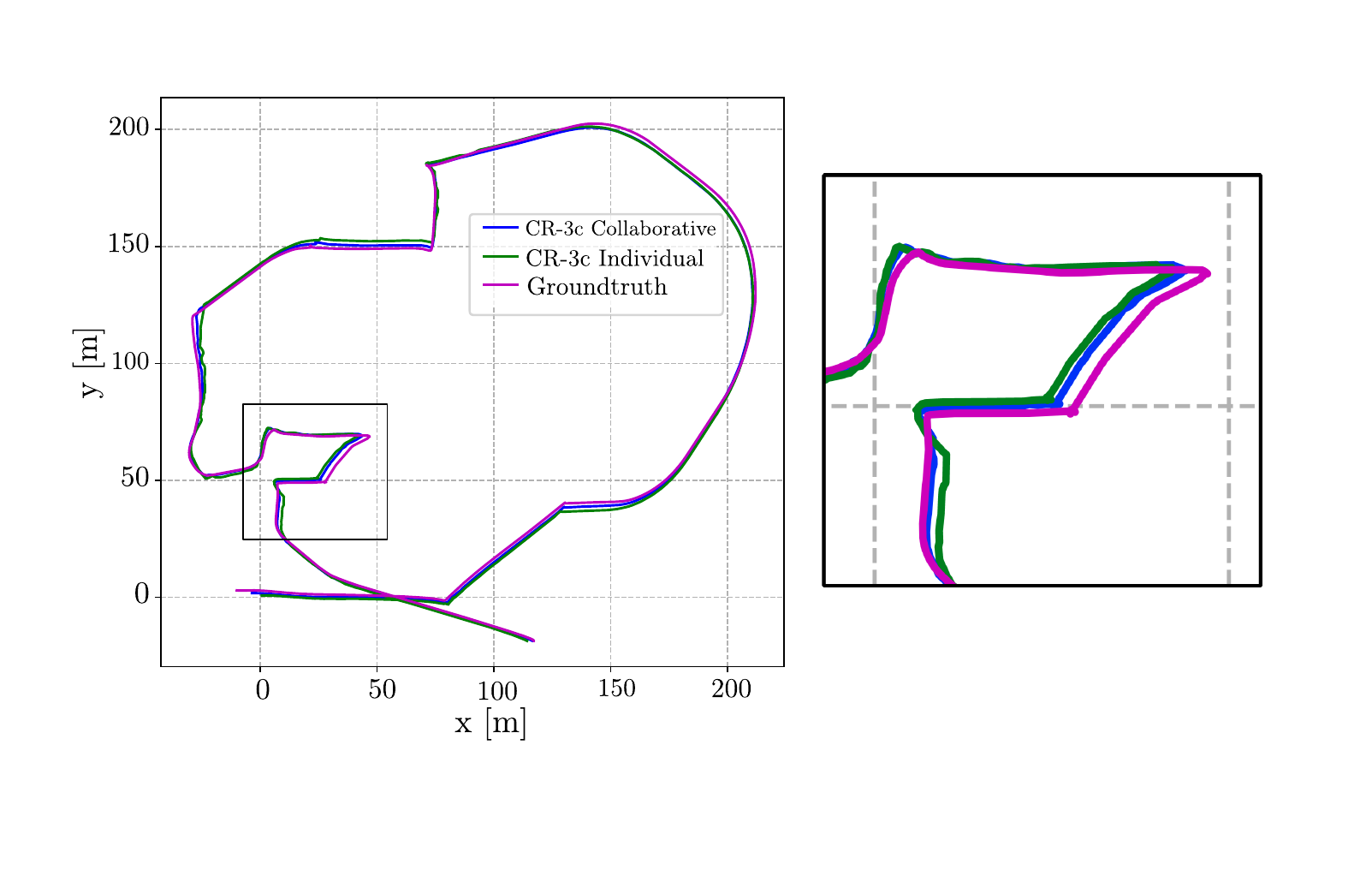}
    \setlength{\abovecaptionskip}{-30pt}
    \caption{The collaborative trajectory and individual trajectory plot with ground-truth for sequence CR-3c (Section \ref{sec:expt2}). The plot has been zoomed on the right for better understanding}
    \label{fig:expt_2_3}   
\end{figure}

We use two of the longest and the most difficult sequences CR-3b and CR-3c, from the CR-Monocular camera dataset for this experiment. We demonstrate the advantages of collaborative SLAM in this experiment. The CR-3b and CR-3c sequences, having been recorded from the same zone, have a high amount of overlap in terms of visible areas. Both the trajectories have a length exceeding 850m and have been recorded at a significant average speed of 2.75-3.75 m/s. A gap of around 10 seconds has been kept between both the agents. The comparison between the individual and collaborative performance has been summarized in Table \ref{tab:expt2}. Here, the collaborative performance refers to the ATE obtained by evaluating the error in poses of the KFs obtained from the server for respective trajectories when both the sequences (CR-3b and CR-3c) were running in parallel. We observe a significant decrease in ATE for the collaborative approach as compared to the individual SLAM. The ATE for CR-3c decreased from 3.49 to 2.21 m, and its comparison in the top view with the ground truth has been made in Figure \ref{fig:expt_2_3}. We also calculate the cumulative error(\%) for the collaborative approach, which has been calculated by applying a single Sim-3 transformation for both the agents. The combined trajectory length comes out to be around 1850 m, and the cumulative error (\%) is observed to be around 0.133, which is less than half of the individual error(\%).

\begin{table}[!h]
\centering
\addtolength{\tabcolsep}{3pt}    
\begin{tabular}{ccccccccc}
\hline
\multirow{2}{*}{\begin{tabular}[c]{@{}c@{}}Ag.\\ No.\end{tabular}} &
  \multirow{2}{*}{\begin{tabular}[c]{@{}c@{}}Seq.\\  Name\end{tabular}} &
  \multicolumn{3}{c}{\multirow{2}{*}{Individual}} &
  \multicolumn{3}{c}{\multirow{2}{*}{Collaborative}} &
  \multirow{2}{*}{Collaborative} \\
  &       & \multicolumn{3}{c}{}   & \multicolumn{3}{c}{}  &                        \\ \hline
 &
   &
  Kfs &
  \begin{tabular}[c]{@{}c@{}}RMSE\\ (m)\end{tabular} &
  \begin{tabular}[c]{@{}c@{}}Error\\ (\%)\end{tabular} &
  Kfs &
  \begin{tabular}[c]{@{}c@{}}RMSE\\ (m)\end{tabular} &
  \begin{tabular}[c]{@{}c@{}}Error\\ (\%)\end{tabular} &
  \begin{tabular}[c]{@{}c@{}}Cumulative \\ Error (\%)\end{tabular} \\ \hline
0 & CR-3b & 528 & 2.3335  & 0.2745 & 471 & 2.1497 & 0.2529 & \multirow{2}{*}{0.133} \\
1 & CR-3c & 568 & 3.49317 & 0.3493 & 522 & 2.2121 & 0.2106 &                        \\ \hline
\end{tabular}
\addtolength{\tabcolsep}{-3pt}
\vspace{2mm}
\caption{Comparison of trajectory error for single-agent and Collaborative SLAM for Section \ref{sec:expt2}}
\label{tab:expt2}
\end{table}

\subsection{Collaborative mapping of CR environment}
\label{sec:expt3}

\begin{figure}[!h]
\setlength{\belowcaptionskip}{-5pt}
     \centering
     \begin{subfigure}[b]{0.47\textwidth}
         \centering
         \includegraphics[width=\textwidth]{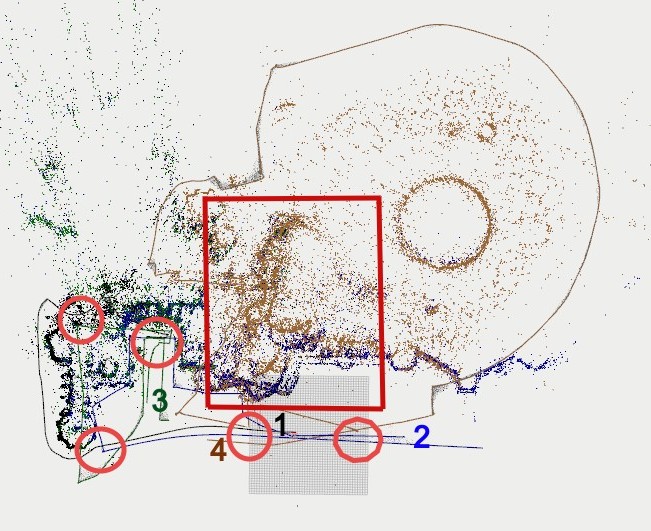}
         \caption{}
         \label{fig:expt3_noGBA}
     \end{subfigure}
     \hfill
     \begin{subfigure}[b]{0.51\textwidth}
         \centering
         \includegraphics[width=\textwidth]{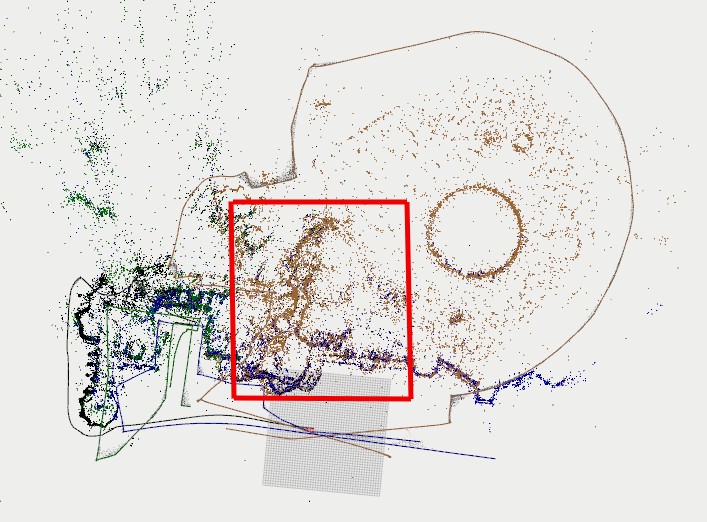}
         \caption{}
         \label{fig:expt3_GBA}
     \end{subfigure}
        \caption{The collaborative map built using 4 agents. The agent trajectories and their corresponding map points are in different colours. It can be seen in the red box in (a) that the brown trajectory has drifted and the map points do not align with the blue map points from other agent. This is corrected by a inter-map loop closure, followed by a GBA which is evident by the overlap in (b). Figure \ref{fig:CR_top} can be used as a top-view reference of the environment.}
        \label{fig:expt3}
\end{figure}

We perform one final experiment in simulation before testing in the real world. We demonstrate the complete mapping of the CR environment using four agents simultaneously. For this, we use one sequence from each of the four different zones of the environment, namely CR-4a, CR-1b, CR-2b, and CR-3b. The agents are deployed in the order given in the above statement with a rough interval of around 10 seconds between two agents. Some of the important places have been marked in Figure \ref{fig:expt3_noGBA}. Here the colored numerals 1, 2, 3, and 4 represent the starting locations of the trajectories of CR-4b, CR-1b, CR-2b, and CR-3b, respectively. The red circles shown in the figure represent the places where either a map-merge or a loop closure occurs. \par

We can see the effect of a global optimization step in Figure  \ref{fig:expt3_noGBA} and Figure \ref{fig:expt3_GBA}. A loop closure has been detected when the agent 3 (brown) visits a place previously mapped by agent 2 (blue). It can be clearly observed how the map points of these two agents have misaligned (marked by a red square) due to the large trajectory of agent 3, which has caused a drift. After the completion of a global optimization step, the aligned map is shown in Figure \ref{fig:expt3_GBA}. The comparison between the individual and collaborative performance has been summarized in Table  \ref{tab:expt3}. The decrease in the ATE, as well as the collaborative error(\%) due to collaborative SLAM, can be clearly seen. It is to be noted that the performance obtained from such a collaborative SLAM is superior to that obtained from multi-session single-agent SLAM where multiple sequences are run one-after the another. This is because the agents are able to receive information from other agents in real-time via the server. Hence, having this additional information acquired from other agents who had previously visited the place helps in improving the performance of the agent’s pose estimate.

\begin{table}[!h]
\centering
\addtolength{\tabcolsep}{3pt}
\begin{tabular}{lllllllll}
\hline
\multicolumn{1}{c}{\multirow{2}{*}{\begin{tabular}[c]{@{}c@{}}Ag.\\ No.\end{tabular}}} &
  \multicolumn{1}{c}{\multirow{2}{*}{\begin{tabular}[c]{@{}c@{}}Seq.\\  Name\end{tabular}}} &
  \multicolumn{3}{c}{\multirow{2}{*}{Individual}} &
  \multicolumn{3}{c}{\multirow{2}{*}{Collaborative}} &
  \multicolumn{1}{c}{\multirow{2}{*}{Collaborative}} \\
\multicolumn{1}{c}{} & \multicolumn{1}{c}{} & \multicolumn{3}{c}{}  & \multicolumn{3}{c}{} & \multicolumn{1}{c}{}    \\ \hline
\multicolumn{1}{c}{} &
  \multicolumn{1}{c}{} &
  \multicolumn{1}{c}{Kfs} &
  \multicolumn{1}{c}{\begin{tabular}[c]{@{}c@{}}RMSE\\ (m)\end{tabular}} &
  \multicolumn{1}{c}{\begin{tabular}[c]{@{}c@{}}Error\\ (\%)\end{tabular}} &
  \multicolumn{1}{c}{Kfs} &
  \multicolumn{1}{c}{\begin{tabular}[c]{@{}c@{}}RMSE\\ (m)\end{tabular}} &
  \multicolumn{1}{c}{\begin{tabular}[c]{@{}c@{}}Error\\ (\%)\end{tabular}} &
  \multicolumn{1}{c}{\begin{tabular}[c]{@{}c@{}}Cumulative \\ Error (\%)\end{tabular}} \\ \hline
0                    & CR-4a                & 382 & 2.101  & 0.677  & 417 & 1.7881 & 0.576 & \multirow{4}{*}{0.1061} \\
1                    & CR-1b                & 313 & 0.9861 & 0.2191 & 290 & 0.5771 & 0.128 &                         \\
2                    & CR-2b                & 222 & 1.168  & 0.3893 & 215 & 0.5843 & 0.194 &                         \\
3                    & CR-3b                & 528 & 2.3335 & 0.2745 & 608 & 2.081  & 0.244 &                         \\ \hline
\end{tabular}
\addtolength{\tabcolsep}{-3pt}
\vspace{2mm}
\caption{Comparison of trajectory error for single-agent and Collaborative SLAM for Section \ref{sec:expt3}}
 \label{tab:expt3}
\end{table}

\subsection{Collaborative mapping of Sadra Fort}
\label{sec:expt4}

\begin{figure}[!h]
\setlength{\belowcaptionskip}{-5pt}
     \centering
     \begin{subfigure}[b]{0.47\textwidth}
         \centering
         \includegraphics[width=\textwidth]{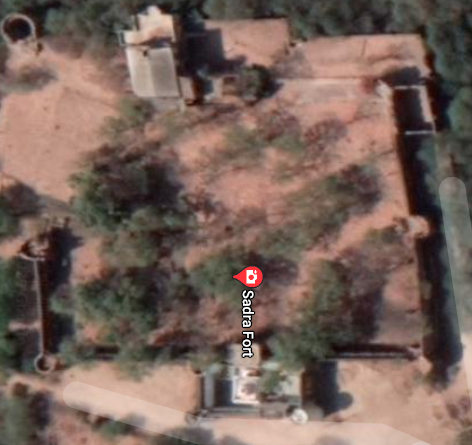}
         \caption{}
         \label{fig:sadra_sat}
     \end{subfigure}
     \hfill
     \begin{subfigure}[b]{0.51\textwidth}
         \centering
         \includegraphics[width=\textwidth]{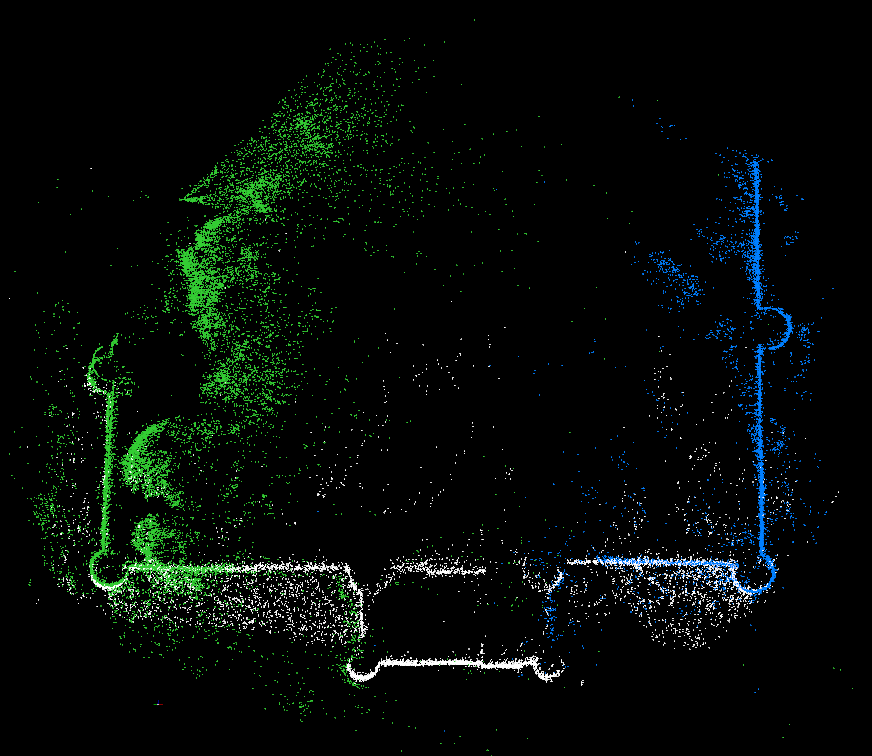}
         \caption{}
         \label{fig:sadra_top}
     \end{subfigure}
        \caption{Results from the real world experiment performed at the Sadra Fort, India. (a) shows the satellite image of the fort, (b) is the top view of the map built by using 3 UAV agents.}
        \label{fig:sadra}
\end{figure}

After validating the approach in simulation, we carry out the experiments on a real-world archaeological site. As discussed in Section \ref{sec:SF-setup-dataset}, we create the SF dataset, where each of the sequences covers a different part of the fort. For this experiment, we choose 3 sequences, namely the SF-2, SF-3, and SF-4. Each of them maps a different wall of the fort, and there are certain overlaps in the trajectories which facilitate the map-fusion and loop closures. We keep all the parameters the same as that in simulation except for the number of ORB features extracted per image. Due to having a higher resolution (1280x720 pixels), we extract 1500 ORB features to ensure proper tracking. The top view of the collaborative map generated can be seen in the Figure \ref{fig:sadra_top}. Here, the green, white, and blue map points represent the map constructed by the agents 0, 1 and 2, respectively. We run the three agents simultaneously with a delay of 10 seconds between them. The agent 0 (green) corresponds to SF-3, agent 1 (white) corresponds to SF-2 and agent 2 (blue) corresponds to SF-4 sequence. Since accurate ground truth is not available for this scenario, we cannot assess the error quantitatively. Though we provide an approximate pose estimate in the dataset, the GPS inaccuracies are too large for evaluating the quantitative performance of the approach. But, we see a great amount of similarity between the satellite image of the fort (Figure \ref{fig:sadra_sat}) and the top view of the map built (Figure \ref{fig:sadra_top}). Moreover, the map points of the same walls from different agents perfectly align with each other indicating a good quality of collaborative estimates.

\subsection{Dense reconstruction of Sadra Fort}
\label{sec:expt5}

A feature-based sparse map is not very useful for archaeological purposes. An accurate texture-based dense map is required for record-keeping as well as for studying architecture and finding defects in the structure. Such a dense map can be highly useful for restoration purposes and also for documentation. Thus, we make attempts at utilizing this sparse map for dense reconstruction using the OpenMVS pipeline \cite{openmvs2020}. The dense reconstruction mainly consists of three parts. To begin with, the sparse point cloud is first converted into a dense point cloud. Next, it has to be converted into a mesh, and as a final step, the texture has to be projected on the mesh. Each of these three functionalities is provided in the OpenMVS pipeline. To generate a dense point cloud, we require the sparse point cloud, undistorted keyframes (images), and the pose of each of the keyframes. Each of the three required inputs are available as an output of the developed collaborative SLAM approach. \par

The standard methods for archaeological mapping involve using an offline SFM approach to generate the sparse point cloud, followed by the dense reconstruction. Though these approaches are very accurate and yield good results, the SFM approaches are generally computationally very expensive and take a lot of time to generate the results. Even a small number of images, such as a thousand images, can take up to a few hours for generating the sparse point cloud, and this time increases exponentially with an increase in the number of images. On the other hand, if we use the collaborative SLAM approach for generating the point cloud, it will take significantly lesser time, and the computation can almost be done online. But, on the downside, it is expected that the accuracy might decrease. This method can find its use in case of rapid mapping requirement or for autonomous mapping missions in which a SLAM system is vital for navigation. \par

\begin{figure}[!h]
\setlength{\belowcaptionskip}{0pt}
     \centering
     \begin{subfigure}[b]{0.52\textwidth}
         \centering
         \includegraphics[width=\textwidth]{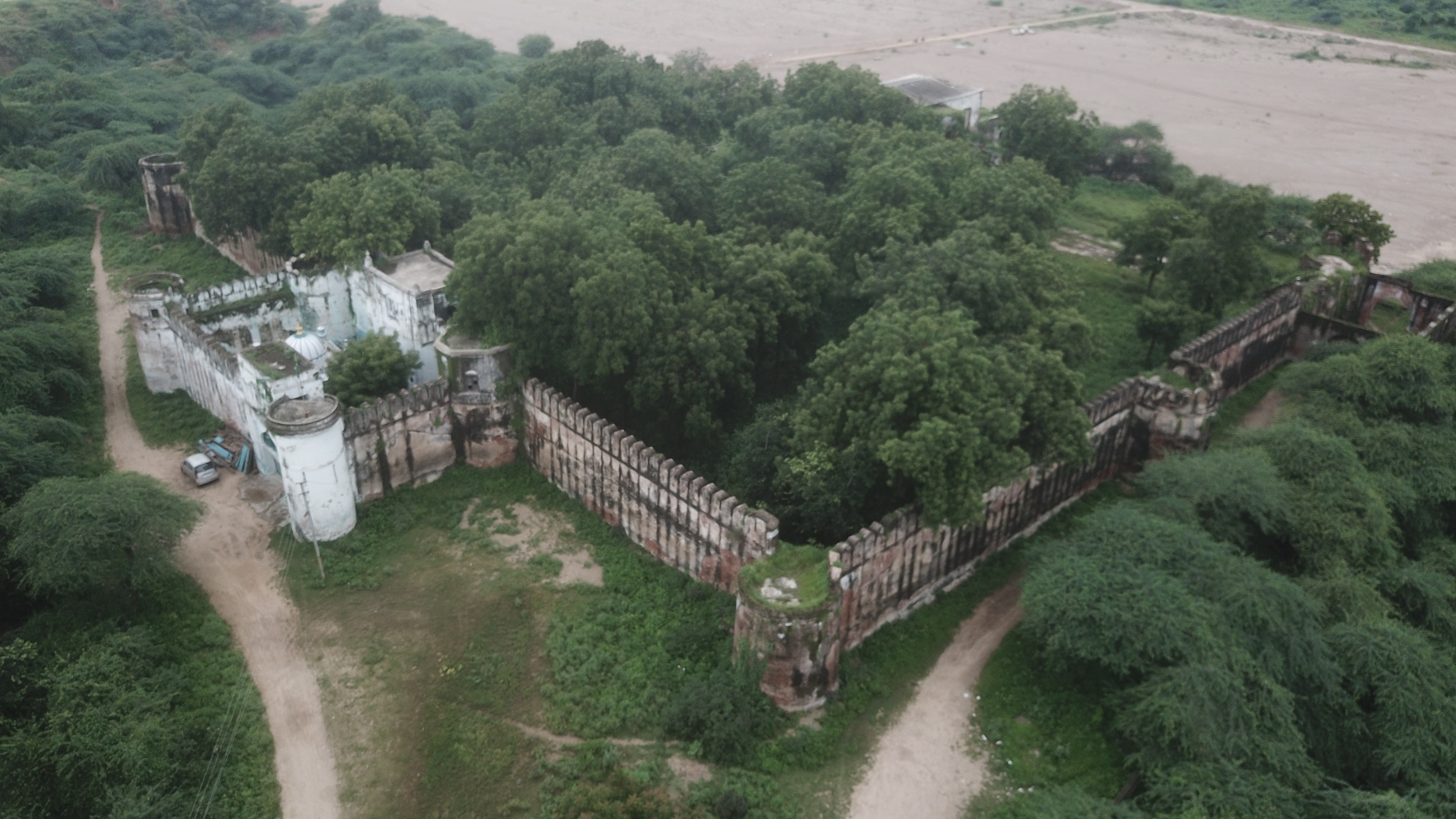}
         \caption{}
         \label{}
     \end{subfigure}
     \hfill
     \begin{subfigure}[b]{0.47\textwidth}
         \centering
         \includegraphics[width=\textwidth]{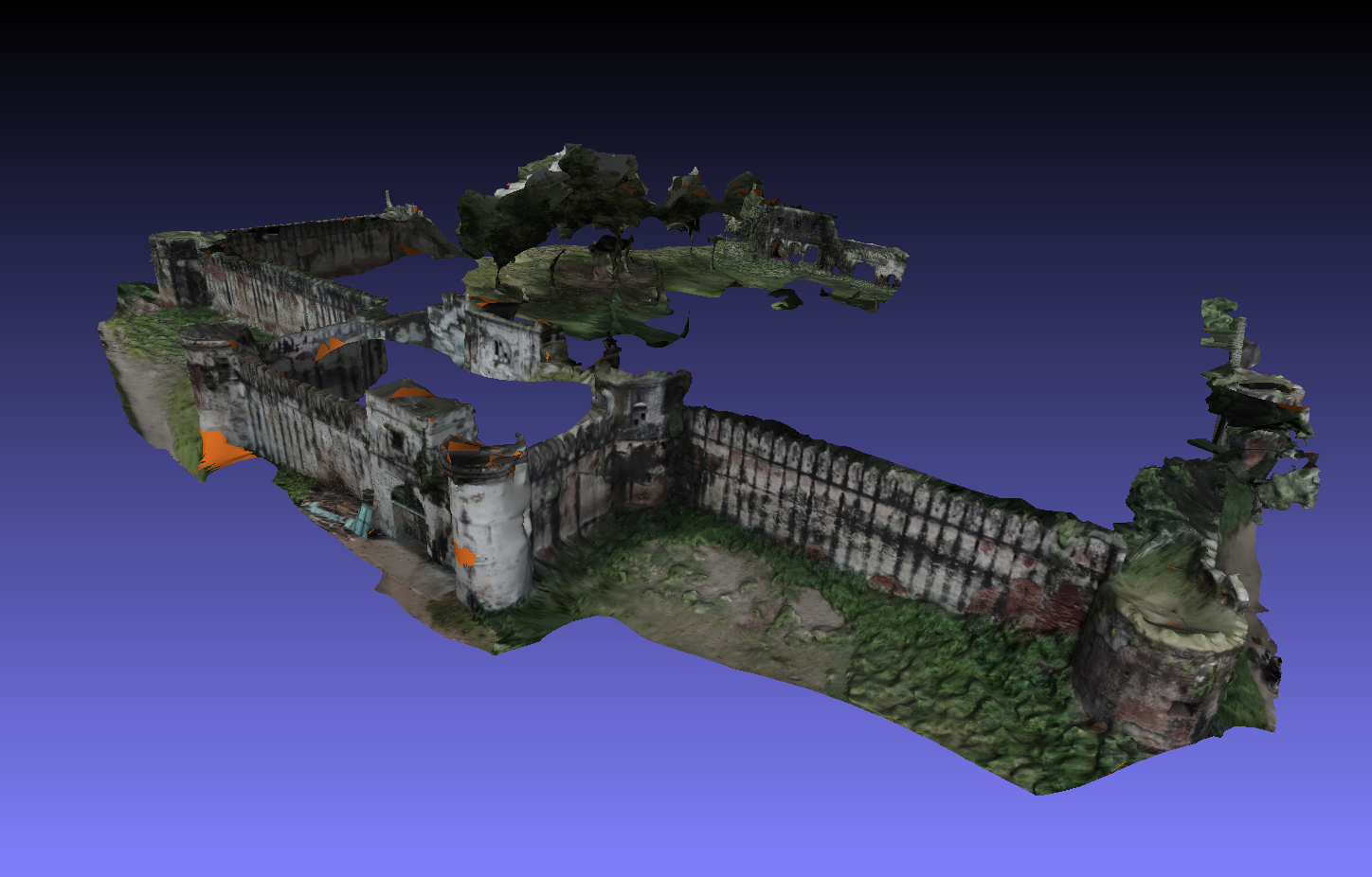}
         \caption{}
         \label{}
     \end{subfigure}
    \hfill
     \begin{subfigure}[b]{0.99\textwidth}
         \centering
         \includegraphics[width=\textwidth]{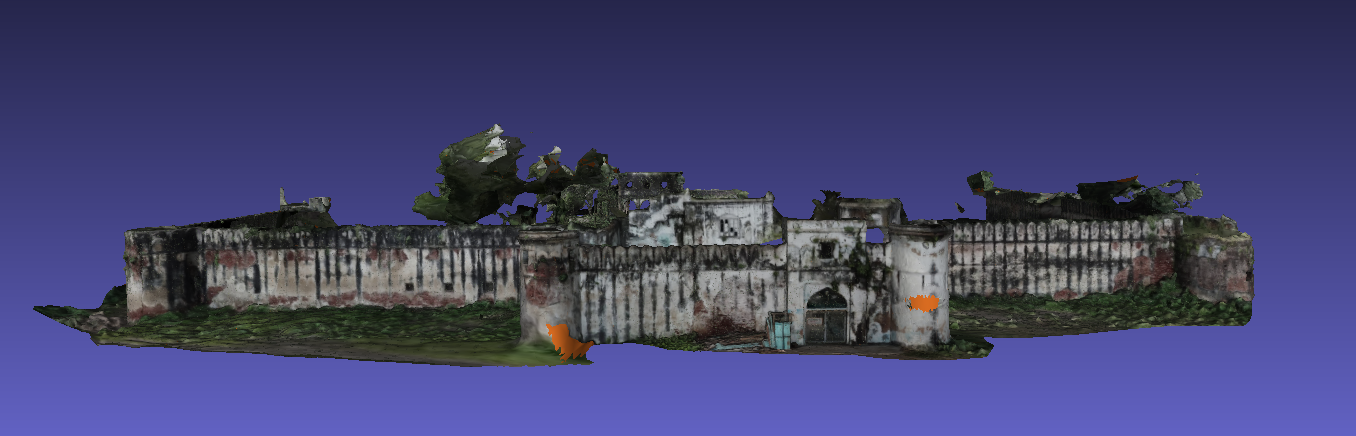}
         \caption{}
         \label{}
     \end{subfigure}
        \caption{An aerial photograph of the Sadra Fort is shown in (a). In (b), the dense model of the Sadra Fort is shown in a similar view as (a). The front view of the fort can be seen in (c).}
        \label{fig:sadra_dense}
\end{figure}

We generate a dense textured model of the Sadra Fort using the OpenMVG + OpenMVS pipeline and the complete SF dataset. OpenMVG is an offline SFM approach that gives the image poses and the sparse map as an output. The obtained results can be seen in Figure \ref{fig:sadra_dense}. We are currently working towards generating results using the output of the collaborative slam approach instead of the openMVG pipeline. This remains one of our primary work to be completed in the future, along with moving in the direction of active collaborative SLAM for autonomous exploration and mapping of sites.

\section{Conclusion and Future Work}

In this paper, we develop a multi-UAV approach for collaborative mapping of archaeological sites. Additionally, we provide two novel datasets to facilitate research on collaborative SLAM algorithms and archaeological mapping. Through experiments in simulation, we demonstrate the advantages of using a collaborative SLAM technique over single-agent SLAM techniques in terms of error and mapping time. We then perform experiments on a 15th-century fort to demonstrate the generation of a sparse collaborative map. This can then be used to generate a dense map using some of the open-source multi-view stereo pipelines. Although using our proposed method for sparse mapping might slightly decrease the accuracy of the dense map, it is beneficial to use when one needs to carry out rapid mapping since it saves a lot of time and computation. Moreover, it can find its application in case of autonomous mapping operations that require an on-board SLAM system. Future directions will focus on developing collaborative active SLAM approaches for autonomous mapping of archaeological sites.

\bibliographystyle{unsrt}  
\bibliography{references} 

\end{document}